\documentclass[final]{nesy2026} 
\usepackage{longtable}
\usepackage{multirow}
\usepackage{float}
\usepackage{booktabs}
\usepackage[load-configurations=version-1]{siunitx} 
\usepackage{needspace}

\newcommand{\specialcell}[2][c]{%
\begin{tabular}[#1]{@{}c@{}}#2\end{tabular}}
\theorembodyfont{\upshape}
\theoremheaderfont{\scshape}
\theorempostheader{:}
\theoremsep{\newline}

\title[Hierarchy-Aware Link Prediction]{Hierarchy-Aware Semantic Losses for Knowledge Graph Link Prediction}

 \author{\Name{Filip Kronström} \Email{filipkro@chalmers.se}\\
 \addr Department of Computer Science and Engineering,\\Chalmers University of Technology and University of Gothenburg, Sweden
 \AND
 \Name{Ross D. King} \Email{rossk@chalmers.se}\\
  \addr Department of Computer Science and Engineering,\\Chalmers University of Technology and University of Gothenburg, Sweden\\
  \addr Department of Chemical Engineering and Biotechnology, University of Cambridge, United Kingdom
 }

\begin{document}

\maketitle

\begin{abstract}
Knowledge graphs are often accompanied by ontological class hierarchies that encode valuable semantic information, yet many link prediction methods either ignore such hierarchies or incorporate them indirectly through additional graph edges. Recent work introduced hierarchy-aware graph neural networks (GNNs), which use semantic losses derived from box embeddings to encourage satisfaction of subclass relationships during GNN-based representation learning. While this approach has shown promise for biological regression tasks, its effectiveness for knowledge graph link prediction has not been investigated.

In this paper we evaluate hierarchy-aware semantic losses on link prediction across three benchmark datasets: AIFB, CoDEx, and BioKG. We combine graph neural network encoders with box-embedding-based semantic losses that encourage learned representations to better satisfy ontology-derived class hierarchies, and compare this approach to both standard link prediction models and models incorporating subclass relations as graph edges. Across all datasets, hierarchy-aware semantic losses significantly improve mean reciprocal rank (MRR) and consistently outperform models that incorporate hierarchy information through additional subclass edges. Relative to the baseline GNN models, MRR improved by 7.6\%, 2.4\%, and 15.5\% on AIFB, CoDEx, and BioKG, respectively. Furthermore, semantic losses consistently outperform the alternative of augmenting the graph with subclass edges.

These results are consistent with ontology-derived class hierarchies providing complementary information to graph structure, and suggest that encouraging hierarchical consistency through semantic losses is an effective and comparatively parameter-efficient mechanism for improving knowledge graph link prediction.
\end{abstract}

\section{Introduction}
\label{sec:intro}

Knowledge graphs (KGs) are widely used to represent structured relational knowledge across domains such as biomedicine, recommender systems, and semantic web applications. Many KGs can be enriched with ontological information, where entities are organised into class hierarchies through relations such as \texttt{subClassOf}. These hierarchies encode valuable background knowledge about conceptual similarity and inheritance structure, and can therefore provide useful inductive bias for representation learning. 

A common task that motivates knowledge graph representation learning is link prediction: estimating missing relations between entities based on the observed graph structure. Contemporary link prediction methods typically learn low-dimensional embeddings of entities and relations by optimising reconstruction objectives over observed triples. Translational methods such as TransE \citep{bordes_translating_2013} and RotatE \citep{sun_rotate_2019}, bilinear models such as DistMult \citep{yang_embedding_2015} and ComplEx \citep{trouillon_complex_2016}, and graph neural network (GNN) approaches such as R-GCNs \citep{schlichtkrull_modeling_2018} have all achieved strong empirical performance on benchmark datasets. However, these methods generally learn representations from the observed graph structure and do not explicitly enforce semantic consistency with ontological hierarchies. As a result, learned embeddings may violate known class inclusion relationships even when such information is available in the data. 

Recent work has explored several approaches for incorporating hierarchical information into representation learning. Within knowledge graph embedding, methods such as HAKE \citep{zhang_learning_2020} and hyperbolic embedding approaches \citep{nickel_poincare_2017} introduce geometric inductive biases that capture hierarchical structure and improve link prediction performance. However, these methods capture hierarchical patterns through geometric properties of the embedding space and do not directly incorporate ontology-derived concepts or logical axioms.

A complementary line of research focuses on embedding ontologies formulated in description logics. Box embeddings \citep{vilnis_probabilistic_2018} provide a geometric representation in which concepts are modelled as axis-aligned hyperrectangles and subsumption relations correspond naturally to geometric containment. This idea has inspired several ontology embedding methods, including ELEm \citep{kulmanov_embeddings_2019}, ELBE \citep{peng_description_2022}, BoxEL \citep{xiong_faithful_2022}, and Box$^2$EL \citep{jackermeier_dual_2024}, which construct geometric representations that approximate model-theoretic interpretations of ontology axioms. These methods show that ontological constraints can be encoded effectively in continuous vector spaces, particularly for biomedical ontologies formulated in lightweight logics such as $\mathcal{EL}^{++}$.

More generally, semantic loss functions provide a mechanism for incorporating symbolic knowledge into neural models by penalising predictions that violate logical constraints during training \citep{xu_semantic_2018}. The geometric representations described above provide a natural way of expressing such constraints in terms of hierarchical relations between concepts.

Building on these ideas, \cite{kronstrom_graph_2026} introduced a hierarchy-aware graph neural network framework that combines GNN-based message passing with semantic loss terms derived from ontological class hierarchies. The method represents classes using box embeddings and optimises semantic losses encouraging subclass containment while simultaneously training on a downstream prediction task. Applied to a heterogeneous biological KG describing the yeast \emph{Saccharomyces cerevisiae}, this approach improved predictive performance on gene deletion fitness prediction and produced embeddings that better reflected the underlying ontology structure.

In this paper, we investigate whether the hierarchy-aware semantic loss framework introduced by \cite{kronstrom_graph_2026} generalises beyond biological regression to knowledge graph link prediction. Our contribution is a systematic empirical evaluation of this framework across three benchmark datasets, together with an analysis of how hierarchy characteristics relate to the observed performance gains. We evaluate the framework on BioKG \citep{walsh_biokg_2020} from the Open Graph Benchmark (OGB) \citep{hu_open_2020}, CoDEx \citep{safavi_codex_2020}, and AIFB \citep{bloehdorn_kernel_2007}, spanning different domains, scales, and degrees of ontological structure.

\vspace{-0.5em}
\section{Hierarchy-aware GNNs}
\vspace{-0.3em}
\label{sec:HGNN}

\begin{figure}
  \centering
  \includegraphics[width=0.6\textwidth]{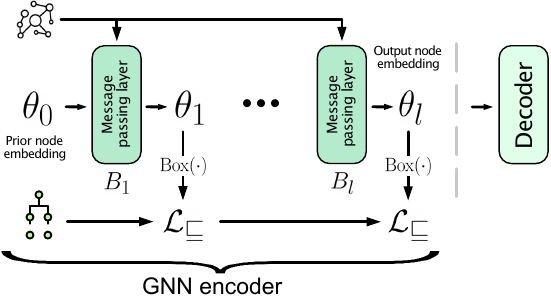}
  \vspace{-1em}
  \caption{Overview of the hierarchy-aware GNN architecture. Semantic losses are applied to the GNN encoder to produce embeddings encouraged to better satisfy subsumptions defined in ontologies. This is optimised jointly with a task-specific decoder, e.g., for link prediction.}
  \label{fig:overview}
  \vspace{-1em}
\end{figure}

Inspired by \cite{kronstrom_graph_2026}, we incorporate semantic losses derived from ontological class hierarchies into GNN-based KG embeddings. The approach combines a GNN encoder with semantic losses derived from class hierarchies. This encoder is learnt along with a task-specific decoder, jointly optimising the semantic losses describing the hierarchical constraints and the task-specific loss. An overview of the architecture can be seen in \figureref{fig:overview}.

We represent concepts as axis-aligned hyperrectangles in a latent space,
\begin{equation}
  \text{Box} = \prod_{i=1}^n[z_i, Z_i],
\end{equation}
where $z_i$ and $Z_i$ denote lower and upper coordinates along dimension $i$. Following \cite{chheda_box_2021}, valid boxes are constructed from latent variables, $\theta$, as
\begin{equation}
  z_i(\theta_i) = \theta_i^z, \qquad Z_i(\theta_i) = z_i + \text{softplus}(\theta_i^Z).
  \label{eq:box_transform}
\end{equation}
Instead of the lower and upper coordinates, boxes can also be represented by their centerpoint, $c_i$, and offset, $o_i$, along each dimension $i$.

Hierarchical relations are modelled geometrically through containment: for a subsumption relation $C\sqsubseteq D$, the box corresponding to $C$ is encouraged to lie inside the box corresponding to $D$. These constraints are incorporated through a distance-based semantic loss operating on box embeddings, inspired by \cite{peng_description_2022,jackermeier_dual_2024} and successfully used as semantic loss by \cite{kronstrom_graph_2026}. For two boxes $C$ and $D$, we define the signed element-wise box distance:
\begin{equation}
  d(C_i,D_i) = |c_i^C - c_i^D| - o_i^C - o_i^D,
  \label{eq:box_distance}
\end{equation}
where negative values indicate overlap between the boxes in a given dimension, while positive values indicate separation. The semantic inclusion loss is then defined as
\begin{equation}
  \mathcal{L}_{\sqsubseteq}(C, D) = \Big|\Big|\Big(\max(0,d(C_i, D_i) + 2o_i^C)\Big)_{i=1}^n\Big|\Big|_2,
  \label{eq:dist_loss}
\end{equation}
which penalises violations of subclass containment. Intuitively, the loss measures how far the child box lies outside the parent box in each dimension. When the child box is fully contained within the parent box, the loss is zero; otherwise it increases in proportion to the extent of the containment violation. \cite{kronstrom_graph_2026} also used a negative loss term to penalise disjoint entities. In this work we omit this term since the taxonomies considered lack such axioms and the link prediction task seems to provide separation between entities. This is further discussed in Appendix~\ref{apd:neg_loss}. The overall semantic loss, denoted $\mathcal{L}_{\sqsubseteq}$, is obtained by summing $\mathcal{L}_{\sqsubseteq}(C,D)$ over all subclass axioms in the hierarchy.

At each GNN layer, node embeddings are transformed into box representations using~\eqref{eq:box_transform}. The semantic inclusion loss is then evaluated over the ontology hierarchy to encourage consistency with the ontology-derived subclass relations. The framework is architecture-agnostic and can be combined with arbitrary downstream objectives. While \cite{kronstrom_graph_2026} combined the semantic loss with an MSE loss for regression, in this work we jointly optimise the binary cross-entropy loss for link prediction and the semantic loss:
\begin{equation}
\mathcal{L} = \mathcal{L}_{\mathrm{BCE}} + \alpha \mathcal{L}_{\sqsubseteq},
\label{eq:combined_loss}
\end{equation}
where $\alpha$ controls the contribution of the semantic loss. Because $\mathcal{L}_{\sqsubseteq}$ is jointly optimised with the link prediction objective, the hierarchy constraints are encouraged rather than guaranteed to hold exactly.

This formulation allows the encoder to exploit both local graph structure and global ontological information during representation learning, while remaining compatible with standard end-to-end optimisation. The decoder is simultaneously optimised for the downstream link prediction task.

\vspace{-0.5em}
\section{Experimental setup}
\vspace{-0.3em}

\subsection{Datasets and hierarchy construction}
\vspace{-0.4em}
\label{sec:data}
We evaluate the proposed hierarchy-aware semantic losses on three benchmark knowledge graph datasets: AIFB \citep{bloehdorn_kernel_2007}, CoDEx \citep{safavi_codex_2020}, and BioKG \citep{walsh_biokg_2020}. The datasets vary substantially in scale, domain, and the availability of hierarchical information, allowing us to investigate the effect of semantic constraints under different conditions. Details about the datasets can be found in \tableref{tab:datasets,tab:biokg}.

\begin{table}[t]
\centering
\caption{Overview of dataset and hierarchy characteristics. $|\mathcal{E}|$ and $|\mathcal{R}|$ denote the number of entities and relations in the original KG. Hierarchy entities and subclass axioms denote the number of additional entities and subclass relations introduced by the hierarchy. Coverage indicates the percentage of KG entities represented in the hierarchy. Domain-specific hierarchy characteristics for BioKG are reported in \tableref{tab:biokg}.}
\vspace{-0.7em}
\label{tab:datasets}
\begin{tabular}{lccccccccc}
\toprule
{\smaller Dataset} & $|\mathcal{E}|$ & $|\mathcal{R}|$ & {\smaller\specialcell{Train\\triples}} & {\smaller\specialcell{Val/Test\\triples}} & {\smaller\specialcell{Hierarchy\\entities}} & {\smaller\specialcell{Subclass\\axioms}} & {\smaller\specialcell{Coverage\\(\%)}} & {\smaller\specialcell{Max\\depth}} & {\smaller\specialcell{Median\\depth}}\\
\midrule
{\smaller AIFB} & {\small2,431} & {\small36} & {\small30,102} & {\small416} & {\small172} & {\small3,890} & {\small100.0} & {\small5} & {\small2}\\
{\smaller CoDEx} & {\small77,951} & {\small69} & {\small551,193} & {\small30,622} & {\small1,986} & {\small69,974} & {\small80.2} & {\small22} & {\small10}\\
{\smaller BioKG}  & {\small93,773} & {\small51} & {\small4,762,678} & {\small162,870} & {\small64,985} & {\small455,560} & {\small84.7} & 58 & 6\\
\bottomrule
\end{tabular}
\end{table}

\begin{table}[t]
\centering
\vspace{-0.5em}
\caption{Hierarchy characteristics for the five domains in BioKG.}
\label{tab:biokg}
\begin{tabular}{lcccccc}
\toprule
Domain & $|\mathcal{E}|$ & \specialcell{Hierarchy\\entities} & \specialcell{Subclass\\axioms} & Coverage (\%) & \specialcell{Max\\depth} & \specialcell{Median leaf\\node depth}\\
\midrule
function & 45,085 & 618 & 93,040 & 100.0 & 16 & 6\\
disease  & 10,687 & 25,804 & 163,606 & 83.4 & 58 & 24\\
sideeffect  & 9,969 & 28,217 & 166,898 & 91.2 & 57 & 16\\
protein   & 17,499 & 2,040 & 21,906 & 82.9 & 6 & 2\\
drug  & 10,533 & 8,306 & 10,110 & 17.1 & 11 & 6\\
\bottomrule
\end{tabular}
\end{table}

\paragraph{AIFB}
A semantic web benchmark containing entities such as persons, projects, organizations, and publications. Since the dataset is derived from a single ontology, class hierarchy information is readily available through explicit \texttt{subClassOf} relations, which are directly used in our experiments. For link prediction, we consider only the \texttt{publication} relation linking persons to publications.

\paragraph{CoDEx}
A multi-domain knowledge graph completion benchmark derived from Wikidata and Wikipedia, covering entities from domains such as writing, entertainment, academia, and science. We use CoDEx-L, the largest variant of the benchmark. Since the dataset does not provide explicit hierarchy information, we construct a class hierarchy using the \texttt{instanceOf} (P31) and \texttt{subClassOf} (P279) properties extracted from Wikidata5M \citep{wang_kepler_2021}.

\paragraph{BioKG}
A heterogeneous biomedical knowledge graph containing entities from five domains: functions, diseases, side-effects, proteins, and drugs. Each domain is associated with a domain-specific class hierarchy. Function entities are directly described by the Gene Ontology (GO) \citep{ashburner_gene_2000}. The protein hierarchy was derived from the HGNC gene family taxonomy \citep{gray_review_2016}, where proteins were assigned to gene families via Entrez Gene identifiers and connected through hierarchical family relations. Drug entities were mapped to the ChemOnt chemical taxonomy using structural classifications obtained through the ClassyFire API \citep{djoumbou_feunang_classyfire_2016}. Disease and side-effect hierarchies were derived from the UMLS Metathesaurus \citep{bodenreider_unified_2004}, using hierarchical relations among concepts originating from multiple biomedical vocabularies, primarily SNOMED CT and MeSH.

BioKG is particularly heterogeneous, both in terms of relation types and entity domains. The graph contains interactions between functions, diseases, side-effects, proteins, and drugs, each associated with its own ontology-derived hierarchy. This provides a useful setting for evaluating whether hierarchy-aware semantic losses can exploit multiple sources of ontological information within a single knowledge graph.

\subsection{Models}
\vspace{-0.4em}

For the GNN encoder we consider both R-GCN \citep{schlichtkrull_modeling_2018} and GraphSAGE \citep{hamilton_inductive_2017} architectures. Preliminary experiments indicated that R-GCN performed best on AIFB, while GraphSAGE was more effective on CoDEx and BioKG. We therefore report results using the strongest encoder for each dataset. This difference is consistent with the contrasting characteristics of the datasets: AIFB is comparatively small and contains a limited number of relation types, whereas CoDEx and BioKG are substantially larger and more heterogeneous.

For BioKG, the graph is represented as a heterogeneous graph with separate node and edge types for each domain and relation. Message passing is implemented using a \texttt{HeteroConv} architecture, where relation-specific GraphSAGE convolutions are applied independently and aggregated by mean pooling. Hierarchy-aware semantic losses are applied separately to the ontology-derived embedding spaces associated with each domain.

The output node embeddings generated by the encoder are provided to a link prediction decoder. We evaluate two decoder types:

\begin{itemize}
  \vspace{-0.5em}
\item \textbf{ComplEx} \citep{trouillon_complex_2016}: a bilinear scoring function commonly used for knowledge graph completion.
\item \textbf{Neural network (NN)}: a small multilayer perceptron operating on concatenated head and tail embeddings. Separate network parameters are learned for each relation type, providing relation-specific decoding while sharing the underlying node representations.
\end{itemize}

Preliminary experiments showed that the ComplEx decoder performed best on AIFB, whereas the neural network decoder was more effective on CoDEx and BioKG. This mirrors the encoder selection: the comparatively small AIFB dataset appears to benefit from the additional inductive bias provided by the structured bilinear scoring function, while the larger datasets benefit from the greater flexibility of the neural network decoder.

To assess the effect of incorporating hierarchy information, we compare three configurations:

\begin{itemize}
    \vspace{-0.5em}
    \item \textbf{Baseline}: original graph without hierarchy information.
    \item \textbf{SC-Edges}: \texttt{subClassOf} relations added directly as graph edges.
    \item \textbf{Semantic loss}: \texttt{subClassOf} relations incorporated through hierarchy-aware semantic losses introduced in Section~\ref{sec:HGNN}.
\end{itemize}

Unlike SC-Edges, the semantic loss approach preserves the original graph topology and instead constrains the learned embedding space. We additionally compare against a conventional ComplEx model without a GNN encoder.

\subsection{Training and evaluation}
\vspace{-0.4em}

CoDEx and BioKG use the standard train, validation, and test splits provided with the benchmarks. Since AIFB does not provide a link prediction split for the \texttt{publication} relation, we randomly partition the corresponding triples into training, validation, and test sets using an 80/10/10 ratio.

Models are trained using binary cross-entropy (BCE) loss for link prediction, with randomly sampled negative examples. For hierarchy-aware models, the semantic loss in \eqref{eq:combined_loss} is jointly optimised with the task-specific loss. Evaluation is based on MRR. For AIFB and CoDEx, each positive validation and test triple is ranked against 1,000 randomly sampled negative triples generated by corrupting the head or tail entity, while BioKG uses the benchmark-provided evaluation negatives.

\vspace{-0.3em}
\section{Results}
\vspace{-0.4em}
\sloppy
We evaluate hierarchy-aware semantic losses on AIFB, CoDEx, and BioKG. The reported results are based on 10 training runs. We compare standard link prediction models, models incorporating subclass relations as graph edges (SC-Edges), and hierarchy-aware semantic losses. Full hyperparameter settings are provided in Appendix~\ref{apd:hyper}.

\begin{table}[h]
  \centering
  \vspace{-0.8em}
\caption{Link prediction performance (MRR) and number of trainable parameters. Results are reported as mean $\pm$ standard deviation over 10 runs. All models trained with the semantic loss significantly outperform the corresponding baseline models (Welch's two-sided t-test, $p<0.05$).}
\vspace{-0.8em}
\label{tab:results}
\small
\begin{tabular}{llcccccc}
\toprule
&
&
\multicolumn{2}{c}{ComplEx}
&
\multicolumn{3}{c}{R-GCN / GraphSAGE} \\
\cmidrule(lr){3-4}
\cmidrule(lr){5-7}
Dataset & Metric
& Baseline & SC-Edges
&
Baseline & SC-Edges & SemLoss \\
\midrule

\multirow{2}{*}{AIFB}
& MRR
& {\small\specialcell[c]{0.4056\\[-0.35em]{\smaller$\pm$0.0085}}}
& {\small\specialcell[c]{0.4336\\[-0.35em]{\smaller$\pm$0.0062}}}
& {\small\specialcell[c]{0.4758\\[-0.35em]{\smaller$\pm$0.0237}}}
& {\small\specialcell[c]{0.4872\\[-0.35em]{\smaller$\pm$0.0257}}}
& {\small\specialcell[c]{\textbf{0.5118}\\[-0.35em]{\smaller$\pm$0.0151}}} \\ [0.5em]
&
\small
\#Params (M)
& 1.32
& 1.35
& 0.51
& 0.51
& 0.51 \\
\midrule

\multirow{2}{*}{CoDEx}
& MRR
& {\small\specialcell[c]{0.3734\\[-0.35em]{\smaller$\pm$0.0010}}}
& {\small\specialcell[c]{0.3700\\[-0.35em]{\smaller$\pm$0.0014}}}
& {\small\specialcell[c]{0.3999\\[-0.35em]{\smaller$\pm$0.0069}}}
& {\small\specialcell[c]{0.4035\\[-0.35em]{\smaller$\pm$0.0074}}}
& {\small\specialcell[c]{\textbf{0.4095}\\{\smaller$\pm$0.0025}}} \\ [0.5em]
&
\small
\#Params (M)
& 39.95
& 40.96
& 6.63
& 6.74
& 6.69 \\
\midrule

\multirow{2}{*}{BioKG}
& MRR
& {\small\specialcell[c]{0.7440\\[-0.35em]{\smaller$\pm$0.0044}}}
& {\small\specialcell[c]{0.6963\\[-0.35em]{\smaller$\pm$0.0223}}}
& {\small\specialcell[c]{0.7232\\[-0.35em]{\smaller$\pm$0.0785}}}
& {\small\specialcell[c]{0.7644\\[-0.35em]{\smaller$\pm$0.0393}}}
& {\small\specialcell[c]{\textbf{0.8354}\\[-0.35em]{\smaller$\pm$0.0144}}} \\ [0.5em]
&
\small
\#Params (M)
& 96.08
& 181.05
& 42.30
& 53.41
& 50.62 \\
\bottomrule
\end{tabular}
\end{table}

\tableref{tab:results} and \tableref{tab:hits_k} in Appendix~\ref{apd:hits} summarise the results on the three benchmark datasets. Across all datasets, the semantic loss models achieved the strongest performance and significantly outperformed both the corresponding baseline and SC-Edges variants (Welch's two-sided t-test, $p<0.05$). The largest improvement was observed on BioKG, where semantic losses increased MRR from 0.723 to 0.835, corresponding to a relative improvement of approximately 15\%. Improvements were also observed on AIFB and CoDEx, with relative gains of 7.6\% and 2.4\%, respectively.

The behaviour of the ComplEx baseline on BioKG is also noteworthy. The absolute performance is somewhat lower than reported in leaderboard evaluations, likely reflecting the smaller model size used here. In addition, augmenting the graph with subclass relations decreases ComplEx performance. This may be due to the large number of hierarchy entities introduced into the graph, which participate primarily in taxonomic relations and substantially increase graph size without directly contributing to the link prediction objective.

As shown in \tableref{tab:results}, the semantic-loss models generally require fewer parameters than the corresponding SC-Edges variants, since hierarchy information is incorporated through the loss function rather than entirely through graph augmentation. While representing hierarchy entities increases model size, the resulting parameter overhead remains smaller than that of SC-Edges and depends on the size of the hierarchy. The increase is negligible for AIFB but more substantial for BioKG, which incorporates several large biomedical ontologies and therefore many more hierarchy entities. Despite using fewer parameters, semantic losses consistently achieved stronger performance than SC-Edges, most notably on BioKG where MRR increases from 0.764 to 0.835 while reducing the parameter count from 53.41M to 50.62M.

The strongest gains were obtained on BioKG, which contains extensive ontology-derived hierarchies spanning multiple biomedical domains. In contrast, the improvements on CoDEx were more modest. This suggests that the effectiveness of semantic regularisation depends on characteristics of the available hierarchy information, an observation we investigate further in the following analyses.

\begin{figure}[h]
  \centering
  \includegraphics[width=\textwidth]{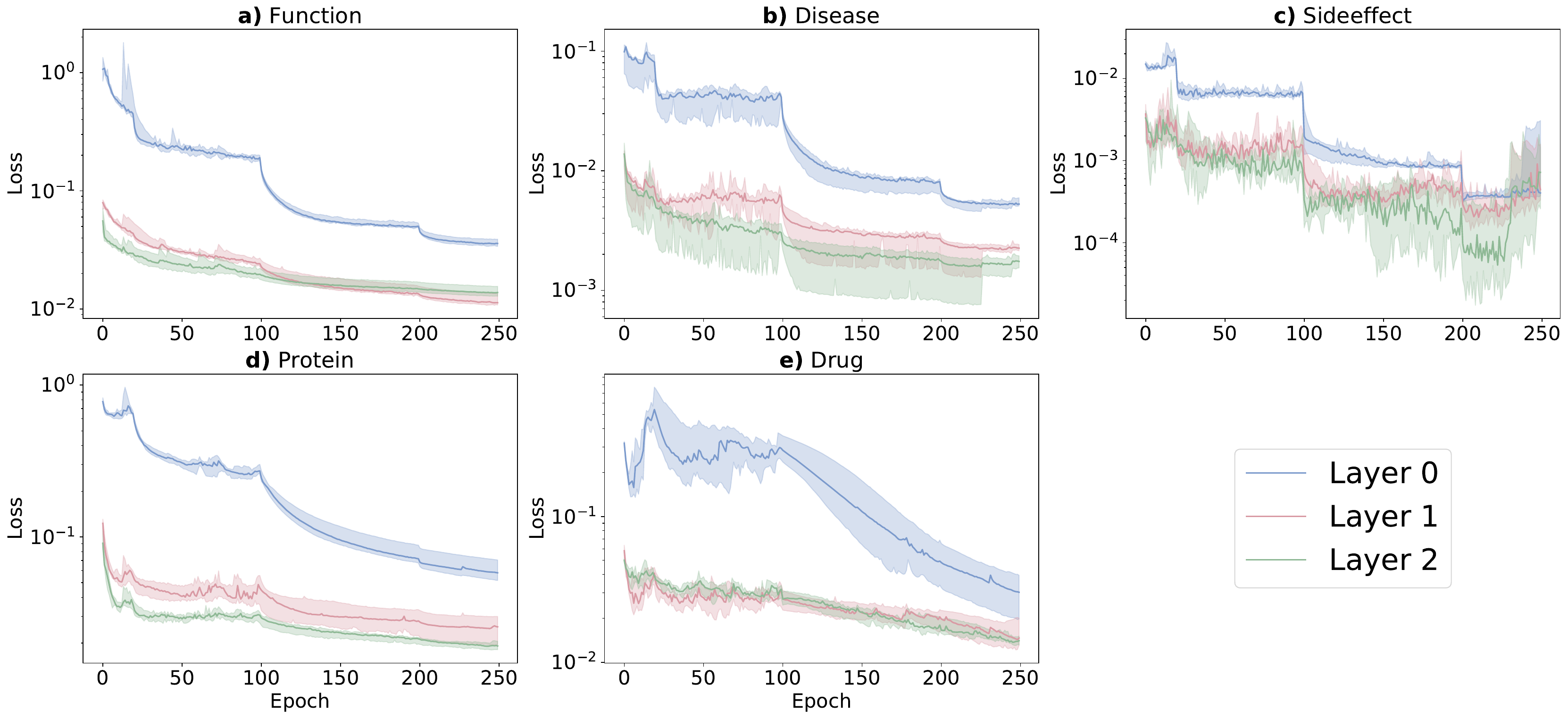}
  \vspace{-2.5em}
  \caption{Evolution of domain-specific semantic losses during training on BioKG. Curves show the median semantic loss over 10 training runs and shaded regions indicate the interquartile range. Losses are reported separately for each ontology domain and GNN layer. The semantic losses decrease throughout training for all domains, indicating increasing satisfaction of ontology-derived hierarchy constraints, although the magnitude of improvement varies between domains. Layer~0 corresponds to the initial node embeddings prior to message passing.}
  \label{fig:biokg-res}
  \vspace{-1em}
\end{figure}

To investigate how the hierarchy constraints are incorporated during training, \figureref{fig:biokg-res} shows the evolution of the semantic losses for the five BioKG ontology domains. The losses decrease consistently throughout training, indicating increasing satisfaction of the ontology-derived hierarchy constraints, although complete satisfaction is not guaranteed because the semantic loss is optimised jointly with the link prediction objective. Reductions are observed across all domains and GNN layers, with the largest decreases occurring for the function, protein, and drug hierarchies. In contrast, the side-effect hierarchy exhibits greater variability and smaller relative improvements. The abrupt reductions observed around epochs 20, 100, and 200 coincide with the scheduled learning-rate decreases described in Appendix~\ref{apd:hyper}.

\begin{figure}[h]
  \centering
  \includegraphics[width=0.9\textwidth]{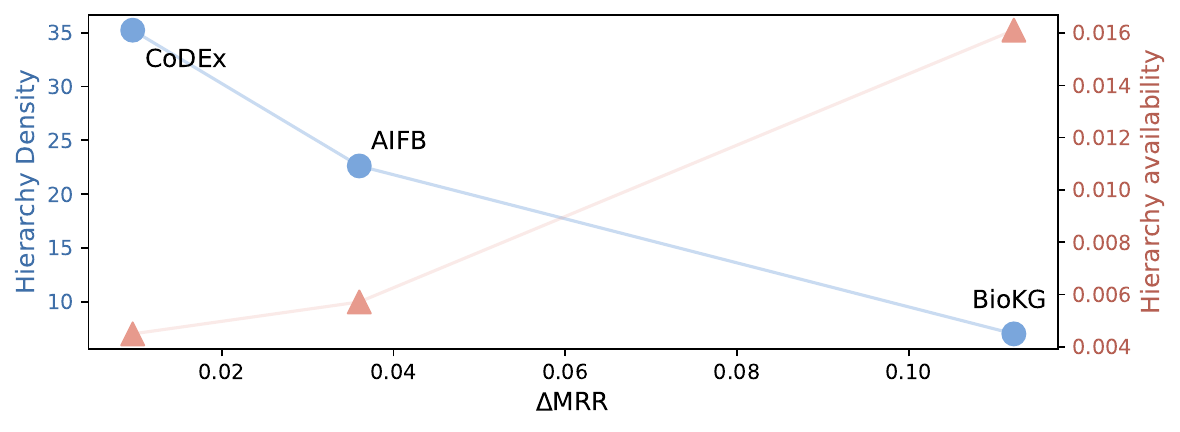}
  \vspace{-1.8em}
  \caption{Exploratory comparison between performance improvements obtained by hierarchy-aware semantic losses and hierarchy characteristics across the three evaluated datasets. The x-axis shows the improvement in mean reciprocal rank ($\Delta$MRR) relative to the corresponding baseline model. \emph{Hierarchy density} is defined as the number of subclass axioms divided by the number of hierarchy entities. \emph{Hierarchy availability} is defined as the number of hierarchy entities divided by the number of hierarchy-covered graph entities and reflects the amount of hierarchy information available per covered graph entity.}
  \label{fig:hierarchy_metrics}
  \vspace{-2em}
\end{figure}

To further investigate the differing improvements across datasets, \figureref{fig:hierarchy_metrics} compares performance gains with two descriptive hierarchy characteristics. We define \emph{hierarchy density} as the number of subclass axioms per hierarchy entity and \emph{hierarchy availability} as the number of hierarchy entities per hierarchy-covered graph entity. Together, these metrics characterise both the local connectivity of the hierarchy and the amount of hierarchy structure available to constrain the learned embeddings. Although only three datasets are available, the figure suggests that larger performance gains are associated with greater hierarchy availability, whereas hierarchy density exhibits the opposite trend.

\vspace{-0.4em}
\section{Discussion}
\vspace{-0.4em}
The results demonstrate that hierarchy-aware semantic losses can improve link prediction performance across a range of knowledge graph domains and scales. Across all three benchmark datasets, semantic losses significantly outperformed both the corresponding baseline models and variants incorporating subclass relations directly as graph edges (SC-Edges). Although the semantic-loss models require additional parameters to represent hierarchy entities, they remain more parameter-efficient than the corresponding SC-Edges variants while achieving stronger predictive performance.

The comparison with SC-Edges is particularly informative because adding subclass relations as additional message-passing edges is a common strategy for incorporating ontology information into graph neural networks. The consistent advantage of semantic losses suggests that explicitly constraining the embedding space to satisfy hierarchical relations may be more effective than relying on message passing alone to propagate taxonomic information. Moreover, semantic losses preserve the original graph topology and avoid introducing large numbers of hierarchy edges into the graph.

Figure~\ref{fig:biokg-res} provides additional insight into how the hierarchy-aware semantic losses evolve during training. Across all five BioKG domains, the semantic losses decrease substantially during training, indicating that the learned embeddings increasingly satisfy the ontology-derived hierarchy constraints. Although reductions are observed across all domains, the side-effect hierarchy exhibits both smaller decreases and lower initial loss values. This suggests that its constraints are already comparatively well satisfied before training, leaving less room for further optimisation.

The magnitude of the improvements varied substantially across datasets. Figure~\ref{fig:hierarchy_metrics} provides an exploratory comparison between performance gains and selected hierarchy characteristics. Although only three datasets are available, the observed trends suggest that performance gains are associated with both higher hierarchy availability and lower hierarchy density. In particular, BioKG exhibits substantially higher hierarchy availability than AIFB and CoDEx while also achieving the largest performance improvements. Conversely, hierarchy density decreases with increasing gains, suggesting that local hierarchy connectivity is not the sole determinant of performance gains.

However, these metrics provide only a simplified view of the hierarchies. The usefulness of hierarchy-derived supervision is likely to depend strongly on how the hierarchies are constructed and how well they capture meaningful semantic relationships. For example, the AIFB hierarchy is derived directly from an ontology, while several BioKG hierarchies originate from curated biomedical ontologies such as the Gene Ontology. In contrast, the CoDEx hierarchy was derived from Wikidata, whose category structure was not primarily designed for ontological reasoning and may therefore provide weaker semantic constraints. Consequently, hierarchy density and hierarchy availability should be viewed as coarse descriptive measures rather than direct indicators of hierarchy quality. Nevertheless, the substantially larger gains observed on BioKG suggest that both the quantity and quality of available hierarchical information influence the effectiveness of hierarchy-aware semantic losses.

Only three datasets were considered in this study, making it difficult to draw strong conclusions regarding which hierarchy characteristics are most important. Future work should evaluate hierarchy-aware semantic losses across a broader range of knowledge graphs and ontology sources, enabling a more systematic investigation of the relationship between hierarchy structure, hierarchy quality, and downstream performance.

More generally, these results indicate that class hierarchies provide complementary information to the relational structure captured by graph neural networks. By encouraging representations to better satisfy ontology-derived hierarchy constraints throughout the encoder, hierarchy-aware semantic losses promote representations that reflect both graph connectivity and conceptual organisation more faithfully. The consistent improvements observed across all datasets suggest that this principle generalises beyond the biological regression task originally studied by \cite{kronstrom_graph_2026} and can be applied effectively to knowledge graph link prediction.

\subsection*{Availability}
\vspace{-0.4em}
Code for generating hierarchies and training models can be found at \url{https://github.com/filipkro/hgnn-lp}

\acks{The computations were enabled by resources provided by the National Academic Infrastructure for Supercomputing in Sweden (NAISS), partially funded by the Swedish Research Council through grant agreement no. 2022-06725. This work was supported by the Wallenberg AI, Autonomous Systems and Software Program (WASP) funded by the Alice Wallenberg Foundation, the UK Engineering and Physical Sciences Research Council (EPSRC) [EP/R022925/2, EP/W004801/1 and EP/X032418/1], and the Chalmers AI Research Centre.}

\bibliography{main}

@article{zhang_learning_2020,
	title = {Learning Hierarchy-Aware Knowledge Graph Embeddings for Link Prediction},
	volume = {34},
	rights = {https://www.aaai.org},
	issn = {2374-3468, 2159-5399},
	url = {https://ojs.aaai.org/index.php/AAAI/article/view/5701},
	doi = {10.1609/aaai.v34i03.5701},
	pages = {3065--3072},
	number = {3},
	journaltitle = {Proceedings of the {AAAI} Conference on Artificial Intelligence},
	shortjournal = {{AAAI}},
	author = {Zhang, Zhanqiu and Cai, Jianyu and Zhang, Yongdong and Wang, Jie},
	urldate = {2026-01-08},
	date = {2020-04-03},
	langid = {english},
}

@inproceedings{xiong_faithful_2022,
	location = {Berlin, Heidelberg},
	title = {Faithful Embeddings for {EL}++ Knowledge Bases},
	isbn = {978-3-031-19432-0},
	url = {https://doi.org/10.1007/978-3-031-19433-7_2},
	doi = {10.1007/978-3-031-19433-7_2},
	pages = {22--38},
	booktitle = {The Semantic Web – {ISWC} 2022: 21st International Semantic Web Conference, Virtual Event, October 23–27, 2022, Proceedings},
	publisher = {Springer-Verlag},
	author = {Xiong, Bo and Potyka, Nico and Tran, Trung-Kien and Nayyeri, Mojtaba and Staab, Steffen},
	urldate = {2026-06-04},
	date = {2022-10-23},
}

@inproceedings{walsh_biokg_2020,
	location = {New York, {NY}, {USA}},
	title = {{BioKG}: A Knowledge Graph for Relational Learning On Biological Data},
	isbn = {978-1-4503-6859-9},
	url = {https://dl.acm.org/doi/10.1145/3340531.3412776},
	doi = {10.1145/3340531.3412776},
	series = {{CIKM} '20},
	shorttitle = {{BioKG}},
	pages = {3173--3180},
	booktitle = {Proceedings of the 29th {ACM} International Conference on Information \& Knowledge Management},
	publisher = {Association for Computing Machinery},
	author = {Walsh, Brian and Mohamed, Sameh K. and Nováček, Vít},
	urldate = {2025-02-17},
	date = {2020-10-19},
}

@inproceedings{hu_open_2020,
	location = {Red Hook, {NY}, {USA}},
	title = {Open graph benchmark: datasets for machine learning on graphs},
	isbn = {978-1-7138-2954-6},
	series = {{NIPS} '20},
	shorttitle = {Open graph benchmark},
	pages = {22118--22133},
	booktitle = {Proceedings of the 34th International Conference on Neural Information Processing Systems},
	publisher = {Curran Associates Inc.},
	author = {Hu, Weihua and Fey, Matthias and Zitnik, Marinka and Dong, Yuxiao and Ren, Hongyu and Liu, Bowen and Catasta, Michele and Leskovec, Jure},
	urldate = {2025-02-17},
	date = {2020-12-06},
}

@inproceedings{bordes_translating_2013,
	location = {Red Hook, {NY}, {USA}},
	title = {Translating embeddings for modeling multi-relational data},
	volume = {2},
	series = {{NIPS}'13},
	pages = {2787--2795},
	booktitle = {Proceedings of the 27th International Conference on Neural Information Processing Systems - Volume 2},
	publisher = {Curran Associates Inc.},
	author = {Bordes, Antoine and Usunier, Nicolas and Garcia-Durán, Alberto and Weston, Jason and Yakhnenko, Oksana},
	urldate = {2025-02-17},
	date = {2013-12-05},
}

@inproceedings{kulmanov_embeddings_2019,
	location = {Macao, China},
	title = {{EL} Embeddings: Geometric Construction of Models for the Description Logic {EL}++},
	isbn = {978-0-9992411-4-1},
	url = {https://www.ijcai.org/proceedings/2019/845},
	doi = {10.24963/ijcai.2019/845},
	shorttitle = {{EL} Embeddings},
	eventtitle = {Twenty-Eighth International Joint Conference on Artificial Intelligence \{{IJCAI}-19\}},
	pages = {6103--6109},
	booktitle = {Proceedings of the Twenty-Eighth International Joint Conference on Artificial Intelligence},
	publisher = {International Joint Conferences on Artificial Intelligence Organization},
	author = {Kulmanov, Maxat and Liu-Wei, Wang and Yan, Yuan and Hoehndorf, Robert},
	urldate = {2025-02-17},
	date = {2019-08},
	langid = {english},
}

@inproceedings{hamilton_inductive_2017,
	location = {Red Hook, {NY}, {USA}},
	title = {Inductive representation learning on large graphs},
	isbn = {978-1-5108-6096-4},
	series = {{NIPS}'17},
	pages = {1025--1035},
	booktitle = {Proceedings of the 31st International Conference on Neural Information Processing Systems},
	publisher = {Curran Associates Inc.},
	author = {Hamilton, William L. and Ying, Rex and Leskovec, Jure},
	urldate = {2025-02-17},
	date = {2017-12-04},
}

@misc{chheda_box_2021,
	title = {Box Embeddings: An open-source library for representation learning using geometric structures},
	url = {http://arxiv.org/abs/2109.04997},
	doi = {10.48550/arXiv.2109.04997},
	shorttitle = {Box Embeddings},
	number = {{arXiv}:2109.04997},
	publisher = {{arXiv}},
	author = {Chheda, Tejas and Goyal, Purujit and Tran, Trang and Patel, Dhruvesh and Boratko, Michael and Dasgupta, Shib Sankar and {McCallum}, Andrew},
	urldate = {2025-02-19},
	date = {2021-09-10},
	langid = {english},
	eprinttype = {arxiv},
	eprint = {2109.04997 [cs]},
}

@inproceedings{jackermeier_dual_2024,
	location = {New York, {NY}, {USA}},
	title = {Dual Box Embeddings for the Description Logic {EL}++},
	isbn = {979-8-4007-0171-9},
	url = {https://dl.acm.org/doi/10.1145/3589334.3645648},
	doi = {10.1145/3589334.3645648},
	series = {{WWW} '24},
	pages = {2250--2258},
	booktitle = {Proceedings of the {ACM} Web Conference 2024},
	publisher = {Association for Computing Machinery},
	author = {Jackermeier, Mathias and Chen, Jiaoyan and Horrocks, Ian},
	urldate = {2025-08-19},
	date = {2024-05-13},
}

@inproceedings{xu_semantic_2018,
	title = {A Semantic Loss Function for Deep Learning with Symbolic Knowledge},
	issn = {2640-3498},
	url = {https://proceedings.mlr.press/v80/xu18h.html},
	eventtitle = {International Conference on Machine Learning},
	pages = {5502--5511},
	booktitle = {Proceedings of the 35th International Conference on Machine Learning},
	publisher = {{PMLR}},
	author = {Xu, Jingyi and Zhang, Zilu and Friedman, Tal and Liang, Yitao and Broeck, Guy},
	urldate = {2025-08-25},
	date = {2018-07-03},
	langid = {english},
}

@inproceedings{vilnis_probabilistic_2018,
	location = {Melbourne, Australia},
	title = {Probabilistic Embedding of Knowledge Graphs with Box Lattice Measures},
	url = {https://aclanthology.org/P18-1025/},
	doi = {10.18653/v1/P18-1025},
	eventtitle = {{ACL} 2018},
	pages = {263--272},
	booktitle = {Proceedings of the 56th Annual Meeting of the Association for Computational Linguistics (Volume 1: Long Papers)},
	publisher = {Association for Computational Linguistics},
	author = {Vilnis, Luke and Li, Xiang and Murty, Shikhar and {McCallum}, Andrew},
	editor = {Gurevych, Iryna and Miyao, Yusuke},
	urldate = {2025-08-27},
	date = {2018-07},
}

@inproceedings{nickel_poincare_2017,
	location = {Red Hook, {NY}, {USA}},
	title = {Poincaré embeddings for learning hierarchical representations},
	isbn = {978-1-5108-6096-4},
	series = {{NIPS}'17},
	pages = {6341--6350},
	booktitle = {Proceedings of the 31st International Conference on Neural Information Processing Systems},
	publisher = {Curran Associates Inc.},
	author = {Nickel, Maximilian and Kiela, Douwe},
	urldate = {2026-01-07},
	date = {2017-12-04},
}

@inproceedings{schlichtkrull_modeling_2018,
	location = {Cham},
	title = {Modeling Relational Data with Graph Convolutional Networks},
	isbn = {978-3-319-93417-4},
	doi = {10.1007/978-3-319-93417-4_38},
	pages = {593--607},
	booktitle = {The Semantic Web},
	publisher = {Springer International Publishing},
	author = {Schlichtkrull, Michael and Kipf, Thomas N. and Bloem, Peter and van den Berg, Rianne and Titov, Ivan and Welling, Max},
	editor = {Gangemi, Aldo and Navigli, Roberto and Vidal, Maria-Esther and Hitzler, Pascal and Troncy, Raphaël and Hollink, Laura and Tordai, Anna and Alam, Mehwish},
	date = {2018},
	langid = {english},
}

@inproceedings{sun_rotate_2019,
	title = {{RotatE}: Knowledge Graph Embedding by Relational Rotation in Complex Space},
	url = {https://openreview.net/forum?id=HkgEQnRqYQ},
	shorttitle = {{RotatE}},
	booktitle = {7th International Conference on Learning Representations, {ICLR} 2019, New Orleans, {LA}, {USA}, May 6-9, 2019},
	publisher = {{OpenReview}.net},
	author = {Sun, Zhiqing and Deng, Zhi-Hong and Nie, Jian-Yun and Tang, Jian},
	urldate = {2026-01-08},
	date = {2019},
}

@inproceedings{trouillon_complex_2016,
	title = {Complex Embeddings for Simple Link Prediction},
	issn = {1938-7228},
	url = {https://proceedings.mlr.press/v48/trouillon16.html},
	eventtitle = {International Conference on Machine Learning},
	pages = {2071--2080},
	booktitle = {Proceedings of The 33rd International Conference on Machine Learning},
	publisher = {{PMLR}},
	author = {Trouillon, Théo and Welbl, Johannes and Riedel, Sebastian and Gaussier, Eric and Bouchard, Guillaume},
	urldate = {2026-01-08},
	date = {2016-06-11},
	langid = {english},
}

@inproceedings{yang_embedding_2015,
	title = {Embedding Entities and Relations for Learning and Inference in Knowledge Bases},
	url = {http://arxiv.org/abs/1412.6575},
	booktitle = {3rd International Conference on Learning Representations, {ICLR} 2015, San Diego, {CA}, {USA}, May 7-9, 2015, Conference Track Proceedings},
	author = {Yang, Bishan and Yih, Wen-tau and He, Xiaodong and Gao, Jianfeng and Deng, Li},
	editor = {Bengio, Yoshua and {LeCun}, Yann},
	urldate = {2026-01-08},
	date = {2015},
}

@inproceedings{kronstrom_graph_2026,
	title = {Graph Neural Network based Hierarchy-Aware Embeddings of Knowledge Graphs: Applications to Yeast Phenotype Prediction},
	doi = {10.48550/arXiv.2605.03690},
	booktitle = {arXiv},
	shorttitle = {Graph Neural Network based Hierarchy-Aware Embeddings of Knowledge Graphs},
	titleaddon = {{arXiv}.org},
	author = {Kronström, Filip and Gower, Alexander H. and Brunnsåker, Daniel and Tiukova, Ievgeniia A. and King, Ross D.},
	date = {2026-05-05},
	langid = {english},
	year = {2026}
}

@inproceedings{safavi_codex_2020,
	location = {Online},
	title = {{CoDEx}: A Comprehensive Knowledge Graph Completion Benchmark},
	doi = {10.18653/v1/2020.emnlp-main.669},
	shorttitle = {{CoDEx}},
	eventtitle = {{EMNLP} 2020},
	pages = {8328--8350},
	booktitle = {Proceedings of the 2020 Conference on Empirical Methods in Natural Language Processing ({EMNLP})},
	publisher = {Association for Computational Linguistics},
	author = {Safavi, Tara and Koutra, Danai},
	editor = {Webber, Bonnie and Cohn, Trevor and He, Yulan and Liu, Yang},
	date = {2020-11},
	year = {2020}
}

@inproceedings{bloehdorn_kernel_2007,
	location = {Berlin, Heidelberg},
	title = {Kernel Methods for Mining Instance Data in Ontologies},
	isbn = {978-3-540-76298-0},
	doi = {10.1007/978-3-540-76298-0_5},
	pages = {58--71},
	booktitle = {The Semantic Web},
	publisher = {Springer},
	author = {Bloehdorn, Stephan and Sure, York},
	editor = {Aberer, Karl and Choi, Key-Sun and Noy, Natasha and Allemang, Dean and Lee, Kyung-Il and Nixon, Lyndon and Golbeck, Jennifer and Mika, Peter and Maynard, Diana and Mizoguchi, Riichiro and Schreiber, Guus and Cudré-Mauroux, Philippe},
	date = {2007},
	langid = {english},
	year = {2007},
}

@misc{peng_description_2022,
	title = {Description Logic {EL}++ Embeddings with Intersectional Closure},
	doi = {10.48550/arXiv.2202.14018},
	number = {{arXiv}:2202.14018},
	publisher = {{arXiv}},
	author = {Peng, Xi and Tang, Zhenwei and Kulmanov, Maxat and Niu, Kexin and Hoehndorf, Robert},
	date = {2022-02-28},
	eprinttype = {arxiv},
	eprint = {2202.14018 [cs]},
	year = {2022}
}

@article{wang_kepler_2021,
	location = {Cambridge, {MA}},
	title = {{KEPLER}: A Unified Model for Knowledge Embedding and Pre-trained Language Representation},
	volume = {9},
	doi = {10.1162/tacl_a_00360},
	shorttitle = {{KEPLER}},
	pages = {176--194},
	journaltitle = {Transactions of the Association for Computational Linguistics},
	journal = {Transactions of the Association for Computational Linguistics},
	publisher = {{MIT} Press},
	author = {Wang, Xiaozhi and Gao, Tianyu and Zhu, Zhaocheng and Zhang, Zhengyan and Liu, Zhiyuan and Li, Juanzi and Tang, Jian},
	editor = {Roark, Brian and Nenkova, Ani},
	date = {2021},
	year = {2021}
}

@article{ashburner_gene_2000,
	title = {Gene Ontology: tool for the unification of biology},
	volume = {25},
	rights = {2000 Nature America Inc.},
	issn = {1546-1718},
	doi = {10.1038/75556},
	shorttitle = {Gene Ontology},
	pages = {25--29},
	number = {1},
	journaltitle = {Nature Genetics},
	journal = {Nature Genetics},
	shortjournal = {Nat Genet},
	publisher = {Nature Publishing Group},
	author = {Ashburner, Michael and Ball, Catherine A. and Blake, Judith A. and Botstein, David and Butler, Heather and Cherry, J. Michael and Davis, Allan P. and Dolinski, Kara and Dwight, Selina S. and Eppig, Janan T. and Harris, Midori A. and Hill, David P. and Issel-Tarver, Laurie and Kasarskis, Andrew and Lewis, Suzanna and Matese, John C. and Richardson, Joel E. and Ringwald, Martin and Rubin, Gerald M. and Sherlock, Gavin},
	year = {2000},
	date = {2000-05},
	langid = {english},
	note = {Number: 1},
}

@article{gray_review_2016,
	title = {A review of the new {HGNC} gene family resource},
	volume = {10},
	issn = {1473-9542},
	doi = {10.1186/s40246-016-0062-6},
	pages = {6},
	journaltitle = {Human Genomics},
	journal = {Human Genomics},
	shortjournal = {Hum Genomics},
	author = {Gray, Kristian A and Seal, Ruth L and Tweedie, Susan and Wright, Mathew W and Bruford, Elspeth A},
	date = {2016-02-03},
	year = {2016}
}

@article{djoumbou_feunang_classyfire_2016,
	title = {{ClassyFire}: automated chemical classification with a comprehensive, computable taxonomy},
	volume = {8},
	issn = {1758-2946},
	doi = {10.1186/s13321-016-0174-y},
	shorttitle = {{ClassyFire}},
	pages = {61},
	number = {1},
	journaltitle = {Journal of Cheminformatics},
	journal = {Journal of Cheminformatics},
	shortjournal = {J Cheminform},
	author = {Djoumbou Feunang, Yannick and Eisner, Roman and Knox, Craig and Chepelev, Leonid and Hastings, Janna and Owen, Gareth and Fahy, Eoin and Steinbeck, Christoph and Subramanian, Shankar and Bolton, Evan and Greiner, Russell and Wishart, David S.},
	year = {2016},
	date = {2016-11-04},
	langid = {english},
}

@article{bodenreider_unified_2004,
	title = {The Unified Medical Language System ({UMLS}): integrating biomedical terminology},
	volume = {32},
	issn = {0305-1048},
	doi = {10.1093/nar/gkh061},
	shorttitle = {The Unified Medical Language System ({UMLS})},
	pages = {D267--D270},
	issue = {Database issue},
	journaltitle = {Nucleic Acids Research},
	journal = {Nucleic Acids Research},
	shortjournal = {Nucleic Acids Res},
	author = {Bodenreider, Olivier},
	date = {2004-01-01},
	year = {2004}
}

\appendix

\Needspace{10\baselineskip}
\section{Negative loss}\label{apd:neg_loss}

In addition to the positive semantic inclusion loss, \cite{kronstrom_graph_2026} employed a negative semantic loss to model concepts that should be kept separate in the embedding space. The primary motivation for this loss is to represent disjointness axioms in ontologies. A secondary benefit is that it discourages degenerate solutions in which many concepts collapse into identical or highly overlapping boxes while still satisfying the positive hierarchy constraints.

The negative loss can be applied either to explicitly disjoint concepts or to randomly sampled negative examples:
\begin{equation}
  \mathcal{L}_{\sqsubseteq}^-(C, D) = \Big|\Big|\Big(\max(0,-d(C_i, D_i))\Big)_{i=1}^n\Big|\Big|\cdot\prod_{i=1}^n\mathbb{I}\big[d(C_i, D_i)<0\big].
  \label{eq:dist_loss_neg}
\end{equation}

Intuitively, the loss penalises overlap between boxes corresponding to concepts that are assumed to be distinct. Combined with the task-specific and positive semantic loss, the overall objective becomes
\begin{equation}
  \mathcal{L} = \mathcal{L}_{\text{TASK}}(y,\hat{y}) + \alpha (\mathcal{L}_{\sqsubseteq} + \beta \mathcal{L}_{\sqsubseteq}^-),
  \label{eq:combined_loss_w_neg}
\end{equation}
where $\beta$ controls the contribution of the negative examples in relation to the positive hierarchy examples.

In the datasets considered in this work, no explicit disjointness axioms are available. Consequently, applying a negative semantic loss would require generating artificial negative examples by randomly sampling pairs of classes. Unlike true disjointness axioms, however, such sampled pairs are not guaranteed to be semantically unrelated. The negative loss may therefore penalise embeddings that correctly place related concepts close together in the latent space, introducing constraints that are not supported by the underlying ontology.

Furthermore, the link prediction objective already provides a strong discriminative signal. To predict relations accurately, the model must learn to distinguish between entities and classes that participate in different parts of the graph, which may reduce the risk of the embedding collapse that motivated the negative loss in the original work. In this setting, the negative loss may provide limited additional information while simultaneously introducing potentially incorrect constraints.

Empirically, preliminary experiments showed no benefit from including the negative loss, and in several cases performance was slightly improved when it was omitted. For these reasons, all experiments reported in this paper use only the positive semantic inclusion loss.

\Needspace{10\baselineskip}
\section{Hyperparameters}\label{apd:hyper}

Hyperparameters were selected based on validation performance. All models were trained using the Adam optimiser with an L2 regularisation weight of $10^{-3}$. Models incorporating semantic loss used a semantic loss weight of $\alpha=10^{-1}$.

AIFB was trained for 120 epochs using a learning rate of $10^{-3}$. CoDEx and BioKG were trained for 150 and 250 epochs, respectively. For these datasets, the initial learning rate was set to $5\times10^{-2}$ and reduced by a factor of two at epochs 10, 40, and 75 for CoDEx, and at epochs 20, 100, and 200 for BioKG. For each run, the model achieving the best validation performance was retained for evaluation on the test set.

Negative examples for link prediction training were generated by uniformly sampling random triples rather than by corrupting heads or tails of positive triples. AIFB used a negative-to-positive ratio of 1000:1 with a positive class weight of 500. CoDEx used a negative-to-positive ratio of 150:1 with a positive class weight of 250, while BioKG used a negative-to-positive ratio of 100:1 with a positive class weight of 200. These settings were selected based on validation performance.

Dataset-specific architectural hyperparameters are summarised in Table~\ref{tab:hyper}.

\begin{table}[ht]
  \centering
  \caption{Hyperparameters for the different datasets.}
  \vspace{1em}
  \label{tab:hyper}
  \begin{tabular}{lcccc}
    Dataset & \specialcell{Embedding\\[-0.35em]dimension} & \specialcell{Hidden\\[-0.35em]dimensions} & \specialcell{R-GCN\\[-0.35em]bases} & \specialcell{NN decoder\\[-0.35em]units} \\
    \toprule
    AIFB & 32 & $[64, 64, 64]$ & 40 & - \\
    CoDEx & 32 & $[64, 128]$ & - & $[32, 8, 1]$ \\
    BioKG & 128 & $[256, 256]$ & - & $[64, 32, 8, 1]$ \\
  \end{tabular}
\end{table}

The baseline ComplEx models were trained using the Adam optimiser with a learning rate of $10^{-3}$. Model selection was performed using early stopping based on validation performance with a patience of 20 epochs. Following preliminary hyperparameter tuning, embedding dimensions of 256 were used for AIFB and CoDEx, while BioKG used an embedding dimension of 512.

\section{Hits@K}
\label{apd:hits}
\begin{table}[H]
\centering
\caption{Link prediction performance (Hits@K). Results are reported as mean $\pm$ standard deviation over 10 runs. $\S$ indicates significant improvement for the model trained with the semantic loss over all baseline models. $\dagger$ and $\ddagger$ indicate significant improvements compared to the ComplEx and ComplEx with \texttt{subClassOf}-edges respectively. * indicates significant improvement compared to the GNN baseline. Improvements are tested with Welch's two-sided t-test ($p<0.05$).}
\label{tab:hits_k}
\small
\begin{tabular}{llcccccc}
\toprule
&
&
\multicolumn{2}{c}{ComplEx}
&
\multicolumn{3}{c}{R-GCN / GraphSAGE} \\
\cmidrule(lr){3-4}
\cmidrule(lr){5-7}
Dataset & Metric
& Baseline & SC-Edges
&
Baseline & SC-Edges & SemLoss \\
\midrule

\multirow{3}{*}[-2.18ex]{AIFB}
& Hits@1
& {\small\specialcell[c]{0.2389\\[-0.35em]{\smaller$\pm$0.0156}}}
& {\small\specialcell[c]{0.2764\\[-0.35em]{\smaller$\pm$0.0113}}}
& {\small\specialcell[c]{0.3478\\[-0.35em]{\smaller$\pm$0.0227}}}
& {\small\specialcell[c]{0.3565\\[-0.35em]{\smaller$\pm$0.0289}}}
& {\small\specialcell[c]{\textbf{0.3861}$^{\S}$\\[-0.35em]{\smaller$\pm$0.0157}}} \\ [0.5em]

& Hits@3
& {\small\specialcell[c]{0.5041\\[-0.35em]{\smaller$\pm$0.0103}}}
& {\small\specialcell[c]{0.5303\\[-0.35em]{\smaller$\pm$0.0106}}}
& {\small\specialcell[c]{0.5334\\[-0.35em]{\smaller$\pm$0.0311}}}
& {\small\specialcell[c]{0.5454\\[-0.35em]{\smaller$\pm$0.0275}}}
& {\small\specialcell[c]{\textbf{0.5671}$^{*\dagger\ddagger}$\\[-0.35em]{\smaller$\pm$0.0190}}} \\ [0.5em]

& Hits@10
& {\small\specialcell[c]{0.7327\\[-0.35em]{\smaller$\pm$0.0113}}}
& {\small\specialcell[c]{0.7334\\[-0.35em]{\smaller$\pm$0.0042}}}
& {\small\specialcell[c]{0.7409\\[-0.35em]{\smaller$\pm$0.0306}}}
& {\small\specialcell[c]{0.7558\\[-0.35em]{\smaller$\pm$0.0289}}}
& {\small\specialcell[c]{\textbf{0.7687}$^{*\dagger\ddagger}$\\[-0.35em]{\smaller$\pm$0.0127}}} \\ [0.5em]
\midrule

\multirow{3}{*}[-2.81ex]{CoDEx}
& Hits@1
& {\small\specialcell[c]{0.3020\\[-0.35em]{\smaller$\pm$0.0019}}}
& {\small\specialcell[c]{0.2971\\[-0.35em]{\smaller$\pm$0.0020}}}
& {\small\specialcell[c]{0.3266\\[-0.35em]{\smaller$\pm$0.0078}}}
& {\small\specialcell[c]{0.3310\\[-0.35em]{\smaller$\pm$0.0082}}}
& {\small\specialcell[c]{\textbf{0.3380}$^{\S}$\\{\smaller$\pm$0.0030}}} \\ [0.5em]
& Hits@3
& {\small\specialcell[c]{0.4111\\[-0.35em]{\smaller$\pm$0.0008}}}
& {\small\specialcell[c]{0.4085\\[-0.35em]{\smaller$\pm$0.0016}}}
& {\small\specialcell[c]{0.4423\\[-0.35em]{\smaller$\pm$0.0066}}}
& {\small\specialcell[c]{0.4448\\[-0.35em]{\smaller$\pm$0.0070}}}
& {\small\specialcell[c]{\textbf{0.4478}$^{\dagger\ddagger}$\\{\smaller$\pm$0.0045}}} \\ [0.5em]

& Hits@10
& {\small\specialcell[c]{0.4914\\[-0.35em]{\smaller$\pm$0.0011}}}
& {\small\specialcell[c]{0.4895\\[-0.35em]{\smaller$\pm$0.0027}}}
& {\small\specialcell[c]{0.5268\\[-0.35em]{\smaller$\pm$0.0064}}}
& {\small\specialcell[c]{0.5274\\[-0.35em]{\smaller$\pm$0.0079}}}
& {\small\specialcell[c]{\textbf{0.5336}$^{*\dagger\ddagger}$\\{\smaller$\pm$0.0050}}} \\ [0.5em]
\midrule
\multirow{3}{*}[-2.18ex]{BioKG}
& Hits@1
& {\small\specialcell[c]{0.6125\\[-0.35em]{\smaller$\pm$0.0060}}}
& {\small\specialcell[c]{0.5563\\[-0.35em]{\smaller$\pm$0.0227}}}
& {\small\specialcell[c]{0.6059\\[-0.35em]{\smaller$\pm$0.0685}}}
& {\small\specialcell[c]{0.6349\\[-0.35em]{\smaller$\pm$0.0524}}}
& {\small\specialcell[c]{\textbf{0.7157}$^{\S}$\\[-0.35em]{\smaller$\pm$0.0302}}} \\ [0.5em]

& Hits@3
& {\small\specialcell[c]{0.8556\\[-0.35em]{\smaller$\pm$0.0036}}}
& {\small\specialcell[c]{0.8187\\[-0.35em]{\smaller$\pm$0.0238}}}
& {\small\specialcell[c]{0.8253\\[-0.35em]{\smaller$\pm$0.0904}}}
& {\small\specialcell[c]{0.8768\\[-0.35em]{\smaller$\pm$0.0348}}}
& {\small\specialcell[c]{\textbf{0.9240}$^{\S}$\\[-0.35em]{\smaller$\pm$0.0184}}} \\ [0.5em]

& Hits@10
& {\small\specialcell[c]{0.9603\\[-0.35em]{\smaller$\pm$0.0017}}}
& {\small\specialcell[c]{0.9167\\[-0.35em]{\smaller$\pm$0.0222}}}
& {\small\specialcell[c]{0.8858\\[-0.35em]{\smaller$\pm$0.1062}}}
& {\small\specialcell[c]{0.9512\\[-0.35em]{\smaller$\pm$0.0240}}}
& {\small\specialcell[c]{\textbf{0.9622}$^{\ddagger}$\\[-0.35em]{\smaller$\pm$0.0239}}} \\ [0.5em]

\bottomrule
\end{tabular}
\end{table}

\end{document}